\RequirePackage{fixltx2e}
\documentclass{ws-ijitdm}
\usepackage[T1]{fontenc}
\usepackage[latin9]{inputenc}
\usepackage{array}
\usepackage{float}
\usepackage{mathtools}
\usepackage{amsmath}
\usepackage{amssymb}
\usepackage{graphicx}
\usepackage[authoryear]{natbib}

\let\today\relax

\makeatletter

\providecommand{\tabularnewline}{\\}
\floatstyle{ruled}
\newfloat{algorithm}{tbp}{loa}
\providecommand{\algorithmname}{Algorithm}
\floatname{algorithm}{\protect\algorithmname}

\let\thanks=\footnote

\usepackage{lmodern}
\usepackage{algorithmicx}
\usepackage{algpseudocode}
\floatname{algorithm}{Function}

\usepackage[pdftex]{hyperref}

\newcommand{\alglinenoNew}[1]{\newcounter{ALG@line@#1}}
\newcommand{\alglinenoPop}[1]{\setcounter{ALG@line}{\value{ALG@line@#1}}}
\newcommand{\alglinenoPush}[1]{\setcounter{ALG@line@#1}{\value{ALG@line}}}

\let\titlefont\undefined
\usepackage{scrextend}

\@ifundefined{showcaptionsetup}{}{%
 \PassOptionsToPackage{caption=false}{subfig}}
\usepackage{subfig}
\makeatother

\begin{document}
\catchline{0}{0}{0000}{}{}
\title{Manipulation of individual judgments in the quantitative pairwise
comparisons method}
\author{Micha\l{} Strada }
\address{Aptiv Services Poland S.A., ul. Powsta?ców Wielkopolskich 13 D-E, 30-707 Kraków, Poland\thanks{e-mail: michal.strada@wp.pl}}
\author{Konrad Ku\l akowski\thanks{corresponding author, e-mail: konrad.kulakowski@agh.edu.pl}}
\address{AGH University of Kraków, Department of Applied Computer Science, al Mickiewicza 30, 30-059 Kraków, Poland}
\maketitle
\begin{abstract}
Decision-making methods very often use the technique of comparing
alternatives in pairs. In this approach, experts are asked to compare
different options, and then a quantitative ranking is created from
the results obtained. It is commonly believed that experts (decision-makers)
are honest in their judgments. In our work, we consider a scenario
in which experts are vulnerable to bribery. For this purpose, we define
a framework that allows us to determine the intended manipulation
and present three algorithms for achieving the intended goal. Analyzing
these algorithms may provide clues to help defend against such attacks.
\end{abstract}

\section{Introduction}

The pairwise comparisons method is the process of comparing entities
in pairs aiming to determine which one is the most preferred \citep{Pan2014arpb}.
Entities very often are referred to as alternatives, while the final
result is in the form of a ranking. The first mention of pairwise
comparisons as a systematic ranking method comes from the 13th century
and is attributed to \emph{Ramon Llull} \citep{Colomer2011rlfa}.
Lull proposed a procedure for selecting candidates based on comparing
them in pairs. This technique might be viewed as in between the electoral
system and a decision-making method in the modern sense. In later
times, pairwise comparisons were used in the context of social choice
and welfare theories \citep{Saari1996tcm,Klamler2003acot,Faliszewski2009lacv},
psychometric measurements \citep{Thurstone1927tmop,Fechner1966eop,Thurstone1994aloc}
and decision-making methods \citep{Mazurek2023aipc}. In 1977, Saaty
published his seminal paper introducing a new decision-making method:
the Analytic Hierarchy Process (AHP) \citep{Saaty1977asmf}. AHP is
based on the quantitative pairwise comparisons of alternatives. As
a result, each of the considered entities is assigned a weight that
determines its importance. Due to its simplicity, but also the presence
of easily available software, AHP has become one of the most popular
decision-making methods. Since AHP was proposed in 1977, many methods
have been developed for calculating rankings for pairwise comparisons
that naturally extended Saaty's approach. Some of them have tried
to find a better alternative to the originally proposed eigenvalue
method (EVM) \citep{Crawford1985anot,Lin2007arff,Bozoki2008sotl,Grzybowski2012noan,Farkas2013arls,Koczkodaj2020oopo,Kazibudzki2022oeop},
others focused on expanding the space of application of the AHP by
defining its various extensions including fuzzy pairwise comparisons
\citep{VanLaarhoven1983afeo,Deng1999mawf,Mikhailov2003dpff,Bartl2022anaf},
interval pairwise comparisons \citep{Saaty1987uaro,Arbel1993psap,Faramondi2023rtrr}
or incomplete pairwise comparisons \citep{Harker1987amoq,Bozoki2010ooco,Faramondi2019iahp,kulakowski2020otgm}.
A separate strand of work involved studies into the properties of
the method \citep{Koczkodaj1996saeo,Bozoki2013aopc,Kulakowski2015otpo,Kazibudzki2016aeop,Csato2020otmo,Kulakowski2021otsb}.
Among many works on this subject, critical voices can also be noticed
\citep{Munier2021ualo}, the most distinctive of which are related
to the rank reversal problem \citep{Belton1983oasc,Saaty1984tlor,Barzilai1994arrn,Munier2021ualo}.

Pairwise comparisons can, however, be found in other decision methods,
such as DEMATEL (direct-influence matrix) \citep{Si2018dtas} and
its different extensions like Q-ROF DEMATEL \citep{Kou2024prof} or
neuro quantum spherical fuzzy (QNSLF) DEMATEL \citep{Kou2023frtr},
WINGS \citep{Michnik2013winl}, ELECTRE, PROMETHEE, MACBETH \citep{Ehrgott2010timc},
BWM \citep{Rezaei2015bwmc} HRE \citep{Kulakowski2014hrea,kulakowski2014hreg},
and others, e.g. the multiple criteria sorting problem \citep{Kadzinski2015mabp}
or weak order approach \citep{Janicki1996awoa}. With these methods,
pairwise comparison becomes a component of all those solutions for
which the above methods are part of the decision-making procedure,
e.g. \citep{Kou2024eomc,Kou2023iedr}.

Comparisons are usually provided by decision-makers or experts. It
is assumed that, acting in their best interest, they try to be accurate
and honest. With this assumption, human errors are the main source
of data inconsistency. Hence, the level of inconsistency determined
by the appropriate index \citep{Brunelli2018aaoi} is usually considered
to be an indicator of the result credibility. Similarly to calculating
the ranking, the initially proposed method of measuring inconsistencies
\citep{Saaty1977asmf} has undergone many modifications and improvements
\citep{Koczkodaj1993ando,Grzybowski2012noan,Kulakowski2014tntb,Aguaron2020ttgc,Kazibudzki2022oeop},
including fuzzy AHP \citep{Ramik2010iopc} and incomplete AHP \citep{Kulakowski2020iifi}.
The properties of various pairwise comparisons matrix inconsistency
indicators have also been analyzed many times \citep{Brunelli2016saso,Koczkodaj2018aoii,Kazibudzki2019aeor}. 

It is easy to imagine, however, that the interest of individual decision-makers
is not the same as that of the organization they represent. This is
the case, for example, when the benefits offered to the decision-makers
come from outside the organization. Most often we deal with bribery,
i.e. a situation when someone unofficially offers certain benefits
to a decision maker (usually money) for taking specific actions in
favor of the payer, but usually against the interest of the organization
the decision-maker acts on behalf of. In order to fulfill the \textquotedbl order\textquotedbl ,
the decision-maker manipulates selected elements of a decision-making
process, influencing the result of the entire procedure in the desired
way. Manipulations are often the subject of research into social choice
and welfare theory. The conducted research concerns methods and mechanisms
of manipulation \citep{Gibbard1973movs,Gardenfors1976mosc,Taylor2005scat,Brandt2016hocs},
susceptibility to manipulation \citep{Kelly1993aasc,Smith1999maoc}
or attempts to protect against manipulation \citep{Faliszewski2010uctp,Faliszewski2009lacv}.
In the literature on electoral systems, we can find an analysis of
selected manipulation methods in relation to various voting methods
\citep{Brandt2016hocs}.

These studies contributed to the increased interest in the problem
of manipulation in decision-making methods. In particular, Yager studied
methods of strategic manipulation of preferential data in the context
of group decision making \citep{Yager2001pspm,Yager2002dasm}. He
proposed modification of the preference aggregation function in such
a way that the attempts of individual agents to manipulate the data
are penalized. The problem with reaching a consensus in the group
of experts can also be seen as a threat to a decision-making process.
Dong et al. \citep{Dong2018cris} proposed a new type of strategic
manipulation related to the relationship of trust between experts
in the group decision-making process. Strategic weight manipulation
supported by analysis of the strategic attribute weight vector design
was considered in \citep{Dong2018swmi}. The rank-reversal property
may also be a potential backdoor facilitating manipulation of the
decision process in AHP. The classic example of rank-reversal in AHP
was given by Belton and Gear \citep{Belton1983oasc}. A constructive
solution to the problem was proposed by Barzilai and Golany \citep{Barzilai1994arrn}.
This phenomenon has been discussed in a number of works, including
\citep{Saaty1984tlor,Dyer1990rota,Schoner1993auat,Fedrizzi2018rrit}.

\section{Problem statement}

\subsection{Models of manipulation\label{subsec:Models-of-manipulations}}

We would like to believe in the honesty of decision-makers. The saying
that ``honesty is the first chapter in the book of wisdom'' attributed
to Thomas Jefferson clearly indicates that honesty is a prerequisite
for making wise decisions. However, can we always count on the honesty
of decision-makers? Probably not. In fact, we do not even expect honesty
from some of them. The observation of Stewart Stafford that \textquotedblleft there
is nothing as pitiful as a politician who is deficient in relaying
untruths\textquotedblright{} suggests that good decision-makers do
not necessarily have to be honest. We know from practice that the
lack of honesty not only concerns politicians.

\begin{figure}[h]
\begin{centering}
\includegraphics[scale=0.45]{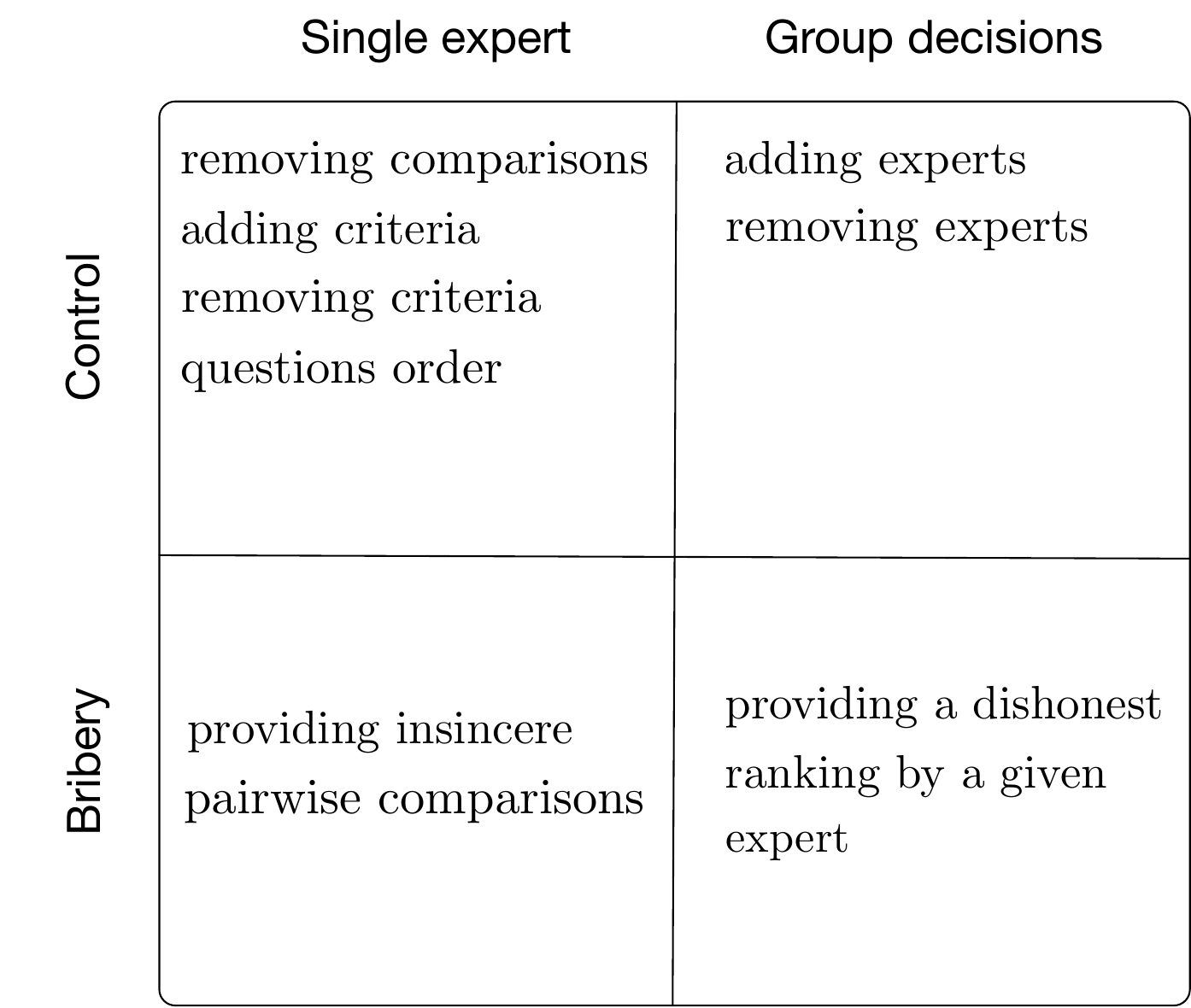}
\par\end{centering}
\caption{Different types of decision data manipulations}
\label{fig:fig_manip}
\end{figure}

In the decision-making process, an unfair attempt to influence the
final outcome of a procedure, hereinafter referred to as manipulation,
can take many forms. The manipulation may be performed by a facilitator,
i.e. a person responsible for supervising the process. He or she may
change the set of criteria or alternatives, decide whether or not
to admit experts, taking these and not other opinions into account
in order to achieve the intended manipulation goal. In the literature
related to social choice theory \citep{Brandt2016hocs}, the facilitator's
manipulations are often referred to as control. Of course, manipulation
can also be committed by the experts (decision-makers)\footnote{For the purposes of this article, we will refer to people who compare
alternatives (who provide their expertise) as experts.}. By expressing a dishonest opinion, they can either strengthen or
weaken one alternative at the expense of others. Since the dishonest
actions of these people are usually associated with corruption, this
type of manipulation is popularly called bribery \citep{Brandt2016hocs}.
Contrary to the theory of social choice and welfare, in decision-making
methods we can consider situations in which there is only one expert.
This naturally introduces an extra dimension to our considerations.
Thus, we can deal with bribery both in group decision making and in
a situation involving only one expert. Similarly, the control may
concern changes in the decision model, both in the case of one and
many experts. Figure \ref{fig:fig_manip} shows the main breakdown
of the types of manipulation according to the person performing the
manipulation and the number of experts involved (affected).

Obviously, this simple diagram cannot describe all possible manipulation
models. In particular, it does not take into account the grafter (bribe
giver), i.e. the person who orders the ranking to be manipulated,
or how the \textquotedbl tampering\textquotedbl{} is paid. In practice,
therefore, in addition to the facilitator and experts, we also need
to consider the actual perpetrator, the grafter who benefits from
the manipulation. It has an indirect influence on the way the decision-making
process is disturbed. For example, they may wish to pay in proportion
to the scale of the manipulation committed. The bribe budget may be
limited, and they may also accept a certain risk of manipulation being
detected, etc. Therefore, when defining the decision-making model
that is the subject of manipulation, we must also take into account
the grafter and the nature of the relationship with the facilitator
and experts.

\subsection{From electoral systems to decision-making methods\label{subsec:From-electoral-systems}}

A grafter can always try to manipulate a decision-making process.
In some cases, due to the nature of the decision-making procedure
and the type of decision data, it may be difficult. For example, Faliszewski
et al. proved that the Llull and Copeland voting systems are resistant
to certain types of manipulation \citep{Faliszewski2009lacv}. In
this case, this resistance is understood in a computational manner,
i.e. a resistant system is one in which the manipulation algorithm
is NP-hard. Much more often, however, manipulation is possible with
the use of less complex and time-consuming procedures. In this case,
it is important to define the conditions under which the algorithm
has a chance to succeed, i.e. enable the intended manipulation and
estimate the cost of the proposed solution. The calculation of the
cost of manipulation requires the acceptance of some payment schema
where the grafter is the payer and the beneficiaries are those involved
in the decision-making process. Manipulation algorithms are often
analyzed in the context of electoral systems \citep{Zuckerman2009aftc,Keller2019nafc,Gupta2021gogc,Aziz2021omov,Schmerler2021svin}.
In such systems, voters play the role of experts, politicians are
the available alternatives, and employees of the election commission
play the role of facilitator. The purpose of such a decision-making
process is to elect politicians for the next term of office.

Apart from the obvious similarities between the quantitative PC (pairwise
comparisons) method and electoral systems (especially the Llull system
\citep{Colomer2011rlfa}), there are many significant differences
and extensions that the former has. First of all, it is quantitative,
i.e. each of the experts determines both the strength of their preferences
and indicates which of the two alternatives is better. As with the
Llull system (but unlike many other electoral systems), experts provide
comparisons for every single pair of options. In the PC method (but
not in the electoral system), the concept of inconsistency is used
to assess the quality of data. In other words, not all data provided
by experts can produce a ranking (some results may be rejected a priori),
while each set of electoral votes leads to a result that must be respected
by society.

\subsection{Algorithms of manipulation}

The above differences, both in terms of the general manipulation model
(Section \ref{subsec:Models-of-manipulations}) and the specificity
of the PC method (Section \ref{subsec:From-electoral-systems}), prompted
us to research the problem of manipulation in the quantitative PC
method used in AHP. In our model, we use the microbribery mechanism
\citep{Faliszewski2009lacv}. This means that the size of the bribe
depends on the degree of manipulation (here the number of manipulated
comparisons). Another limitation that the model must take into account
is the inconsistency of the comparisons set. This means that the manipulation,
in order to be effective, cannot significantly disturb the inconsistency
of the set of comparisons.

Based on these assumptions, we defined two heuristic manipulation
algorithms. Each of the algorithms tries to propose the smallest possible
number of micro-bribes so that the result meets the given level of
data consistency. The tests performed prove that the algorithms do
not always succeed.

Our motivation behind formulating the algorithms outlined in this
paper stemmed from the micro-bribery techniques discussed in the research
by Faliszewski \citep{Faliszewski2009lacv} Consequently, we decided
to develop computational approaches for promoting specific alternatives.
Our primary emphasis was placed on a particular form of micro-bribery,
where the primary concern was minimizing the number of changed comparisons
required to affect the ranking.

\section{Pairwise comparisons}

In the PC method, an expert, often referred to as a decision maker
(DM), compares alternatives in pairs. It is convenient to represent
the results of the comparisons in the form of an $n\times n$ matrix

\begin{equation}
\boldsymbol{C}=\left[c_{ij}\right],\label{eq:Input and output of the PV algorithm}
\end{equation}
where $n$ denotes the number of alternatives, and every single element
$c_{ij}$ indicates the result of comparing between the $a_{i}$ and
$a_{j}$ alternatives. It is natural to expect that if comparing alternative
$a_{i}$ to $a_{j}$ makes $c_{ij}$, then the reverse comparison
$a_{j}$ to $a_{i}$ gives

\begin{equation}
c_{ij}=\frac{1}{c_{ji}}.\label{eq:Reciprocity requirement}
\end{equation}
If it is so for every $i,j\in\{1,\ldots,n\}$, then the PC matrix
$\boldsymbol{C}$ is said to be reciprocal. 

The purpose of the PC method is to calculate a ranking of $n$ alternatives
$A=\{a_{1},\ldots,a_{n}\}$ using a PC matrix $C$ as the input. We
will write down the calculated ranking value of $a_{i}$ as $w(a_{i})$.
The complete ranking takes the form of a priority vector: 
\[
w^{T}=\left[w(a_{1}),\ldots,w(a_{n})\right]^{T}.
\]
Very often it is assumed that the priority vector is rescaled so the
sum of its entries equals $1$, i.e. $\sum_{i=1}^{n}w(a_{i})=1$.
This assumption enables an easy assessment of the significance of
a given alternative without having to consider other priorities.

There are many different ways to compute a priority vector using PC
matrices. One of the most popular is the Eigenvalue method (EVM) proposed
by Saaty in his seminal work on AHP \citep{Saaty1977asmf}. In this
approach, the priority vector is calculated as an appropriately rescaled
principal eigenvector of $C$. Hence, the priority vector $w=\left[w(a_{1}),\ldots,w(a_{n})\right]^{T}$
is calculated as 
\begin{equation}
w(a_{i})=\frac{\widehat{w}_{i}}{\sum_{j=1}^{n}\widehat{w}_{j}},\label{eq:evm-element-val}
\end{equation}
where $\widehat{w}$ and $\lambda_{\textit{max}}$ are the principal
eigenvector and the principal eigenvalue of $C$ correspondingly,
i.e. they satisfy the equation

\begin{equation}
\boldsymbol{C}\widehat{w}=\lambda_{\textit{max}}\widehat{w}.\label{eq:eigenvalueeq}
\end{equation}
The second popular priority deriving procedure is the geometric mean
method (GMM) \citep{Crawford1985taos}. According to this approach,
the priority of the i-th alternative is determined by the geometric
mean of the i-th row of a PC matrix. Thus, after rescaling, the priority
vector is given as $w=\left[w(a_{1}),\ldots,w(a_{n})\right]^{T}$
where 
\[
w(a_{i})=\frac{\left(\prod_{j=1}^{n}c_{ij}\right)^{1/n}}{\sum_{k=1}^{n}\left(\prod_{j=1}^{n}c_{kj}\right)^{1/n}}.
\]

Since elements of $\boldsymbol{C}$ correspond to mutual comparisons
of alternatives, one can suppose that they correspond to the ratios
formed by the ranking results for individual alternatives, i.e. 
\[
c_{ij}=\frac{w(a_{i})}{w(a_{j})},\,\,\,\text{for}\,\,i,j=1,\ldots,n.
\]
This entails the expectation of transitivity between the results of
comparisons, according to which, for every three alternatives $a_{i},a_{k}$
and $a_{j}$, it holds that
\[
\frac{w(a_{i})}{w(a_{k})}\cdot\frac{w(a_{k})}{w(a_{j})}=\frac{w(a_{i})}{w(a_{j})},
\]
i.e.,

\begin{equation}
c_{ik}c_{kj}=c_{ij}.\label{eq:entries_in_C}
\end{equation}
The matrix $C$ where (\ref{eq:entries_in_C}) holds is called consistent.
Unfortunately, in practice, PC matrices are very often inconsistent.
Thus, for some number of alternatives (\ref{eq:entries_in_C}) does
not hold. This also means that $i,j$ exist for which
\[
c_{ij}\neq\frac{w(a_{i})}{w(a_{j})}.
\]
In addition to detecting the fact that a given matrix is inconsistent,
it is also possible to determine the degree of its inconsistency.
For this purpose, inconsistency indices are used \citep{Brunelli2018aaoi}.

One of the most frequently used indices is the consistency index $\textit{CI}$
\citep{Saaty1977asmf} defined as

\begin{equation}
\textit{CI}\left(\boldsymbol{C}\right)=\frac{\lambda_{\textit{max}}-n}{n-1}.\label{eq:saaty-idx}
\end{equation}
In addition to $\textit{CI},$ Saaty introduced the consistency ratio
$\textit{CR}$ defined as 
\[
\textit{CR}\left(\boldsymbol{C}\right)=\frac{\textit{CI}(C)}{\textit{RI}(C)},
\]
where $\textit{RI}(\boldsymbol{C})$ is the average inconsistency
for the fully random PC matrix with the same dimensions as $\boldsymbol{C}$.
According to Saaty's proposal, every PC matrix $\boldsymbol{C}$ for
which $\textit{CR}\left(\boldsymbol{C}\right)>0.1$ is considered
as too inconsistent to be the basis for the ranking. As this threshold
was chosen arbitrarily, some researchers question its value \citep{BanaeCosta2008acao,Karapetrovic1999aqca}.

Another inconsistency index used to determine inconsistency is the
Geometric Consistency Index proposed by Crawford and Williams \citep{Crawford1985anot}
and then standardized by Aguaron and Moreno-Jimenez \citep{Aguaron2003tgci},
defined as
\begin{equation}
\textit{GCI}\left(\boldsymbol{C}\right)=\frac{2}{(n-2)(n-1)}\sum_{i<j}\ln^{2}\left(c_{ij}\frac{w(a_{j})}{w(a_{i})}\right),\label{eq:gci-idx}
\end{equation}
where $w$ is derived with use of the geometric mean method. There
is still a question of the threshold value of $GCI\left(\boldsymbol{C}\right)$
below which the matrix $\boldsymbol{C}$ will be considered sufficiently
consistent \citep{Aguaron2003tgci}. Following the approach proposed
by Saaty \citep{Saaty1994htma} $5\%$ for $n=3$, $8\%$ for $n=4$
and $10\%$ for $n>4$, threshold values can be calculated empirically.

We can also calculate average inconsistency using the Index of Square
Logarithm Deviations proposed by Kazibudzki \citep{Kazibudzki2022oeop}.
This index is defined as
\begin{equation}
\textit{ISLD}\left(\boldsymbol{C}\right)=median\left\{ \ln^{2}\sum_{j=1}^{n}\left(c_{ij}\frac{w(a_{j})}{nw(a_{i})}\right)\right\} _{i=1,\ldots,n}.\label{eq:kazib-idx}
\end{equation}

All of the indices mentioned above take into account every pairwise
comparison \citep{Mazurek2018snot}, so they give us a good insight
how inconsistent is the matrix as a ``whole'', unlike other indices
which rely on local irregularities (like the Koczkodaj's Index) \citep{Koczkodaj1993ando}.

\section{Decision model}

For the purposes of this study, we assumed that the person performing
the manipulation is an expert, and his/her (or grafter's) goal is
such a modification of the given PC matrix that the ranking of the
selected alternative $a_{p}$ will become higher than that of the
selected alternative $a_{q}$. 

Formally speaking, let $r_{C}$ be a function mapping alternatives
to their ranking positions calculated using a certain ranking algorithm\footnote{We tested the two most popular EVM and GMM methods in the conducted
experiments. However, one can use any well-behaved (possibly monotone-behaved)
method. We do not require strict monotony. For example, although Csató
and Petróczy showed that EVM is not monotonic, i.e., increasing $c_{ij}$
does not always increase $w(a_{i})$, the Monte Carlo tests proved
this phenomenon is minimal \citep{Csato2020otmo}. Hence, in practice,
we can neglect this fact for not too inconsistent PC matrices.\label{fn:monotony-remark}} and a given matrix $C$, 
\[
r_{C}:\{a_{1},\ldots a_{n}\}\rightarrow\{1,\ldots,n\}.
\]
In other words, it holds that $r_{C}(a_{i})<r_{C}(a_{j})$ if and
only if $w(a_{i})<w(a_{j})$ for each $i,j\in\{1,\ldots,n\}$, $i\neq j$. 

Let $1\leq p<q\leq n$ be selected indices of alternatives. The purpose
of the designed algorithms is to transform $C$ into $C'$ so that
if $r_{C}(a_{p})\leq r_{C}(a_{q})$ then $r_{C'}(a_{q})<r_{C'}(a_{p})$.
The ability to change individual comparisons implies that the expert,
and thus the grafter, knows how the ``real'' set of comparisons
should look. They must also be able to communicate with each other
to determine how many comparisons can be modified.

According to the adopted micro-bribery strategy, the cost of manipulation
depends on the number of changed comparisons. In our model, the cost
of changing one comparison is fixed and amounts to 1 unit and does
not depend on the size of the change. The adoption of such an assumption
prefers a manipulation with the fewest possible number of modified
comparisons. 

Although in the original AHP the relative strength of preferences
is expressed on a fundamental scale, i.e. for a PC matrix $C=[c_{ij}]$
the value $c_{ij}\in\{1/9,1/8,\ldots,1/2,1,2,\ldots,9\}$, we assume
that the result of individual comparisons is a positive real number
within the range $[1/9,9]$. On the one hand, such an assumption simplifies
the reasoning, but on the other does not limit the proposed solution,
i.e. the resulting algorithms can be easily extended to use any discrete
scale.

The created algorithms must be inconsistency aware. Therefore, they
should try to make the inconsistency of the final result less than
the value set in advance (in AHP it is commonly assumed that the acceptable
inconsistency of $C$ needs to be less than $0.1$, i.e. $\textit{CR}(C)<0.1$
which in most cases also implies that $\textit{CI}(C)<0.1$ \citep{Saaty1977asmf}).
The created algorithms should be able to use various indices (like
the \emph{GCI}, \emph{CI} or \emph{ISLD}), therefore, the threshold
value must depend on the index used.

\section{Row algorithm\label{sec:Row-algorithm}}

The first manipulation algorithm is based on the observation that,
in general\footref{fn:monotony-remark}, increasing $c_{ij}$ causes
increasing $w(a_{i})$ at the expense of all other priority values.
Thus, providing that our goal is to boost alternative $a_{p}$ so
that $r(a_{p})>r(a_{q})$, we will try to increase the subsequent
elements in the p-th row of $C$ with inconsistency of the PC matrix
in mind.

Assuming that $r$ is monotone, to boost the weight of the i-th alternative
let us first try to amplify the i-th row of $C$. Since $r_{C}(a_{p})\leq r_{C}(a_{q})$
then let the first modified element be $c_{pq}$. As $c_{pq}$ is
the direct comparison of the pair $(a_{p},a_{q})$ we suppose that
it may have the greatest impact on their relative position.

Let $\alpha$ be a positive real number such that $\alpha>\max\{1,c_{pq}\}$.
Without loss of generality, we may assume that $q=1$ and $p=n$ .
Thus, in the first step of the algorithm, we update $c_{qp}$ and
$c_{pq}$ i.e. $c_{n1}$ and $c_{1n}$ by $\alpha$ and $1/c_{n1}$
correspondingly. Thus, the first approximation of $C$, denoted as
$\boldsymbol{C}^{\left(1\right)}$, is given as:

\[
\boldsymbol{C}=\boldsymbol{C}^{\left(0\right)}=\begin{bmatrix}1 & c_{12} & \cdots & c_{1n}\\
c_{21} & 1 &  & c_{2n}\\
\vdots &  & \ddots & \vdots\\
c_{n1} & c_{n2} & \cdots & 1
\end{bmatrix}^{\left(0\right)}\xrightarrow[1^{\textit{st}}\,\,\textsl{step}]{\textsl{}}
\]

\[
\xrightarrow[1^{\textit{st}}\,\,\textsl{step}]{\textsl{}}\begin{bmatrix}1 & c_{12} & \cdots & 1/\alpha\\
c_{21} & 1 &  & c_{2n}\\
\vdots &  & \ddots & \vdots\\
\alpha & c_{n2} & \cdots & 1
\end{bmatrix}^{\left(1\right)}=\boldsymbol{C}^{\left(1\right)}
\]

If we find $\alpha$ such that $r_{C^{(1)}}(a_{q})<r_{C^{(1)}}(a_{p})$
then the algorithm stops. Otherwise, we may request further modifications.
Since

\begin{equation}
c_{pj}=c_{pq}c_{qj},\label{eq:Matrix C elements dependence}
\end{equation}
one assumes, that the next element to change is $c_{n2}$, thus $c_{n2}=c_{n1}c_{12}=\alpha c_{12}$.
So, the update for the pair $\left(c_{n2},c_{2n}\right)$ is $\left(\alpha c_{12},1/\alpha c_{12}\right)$.

\[
\boldsymbol{C}^{\left(1\right)}=\begin{bmatrix}1 & c_{12} & \cdots & 1/\alpha\\
c_{21} & 1 &  & c_{2n}\\
\vdots &  & \ddots & \vdots\\
\alpha & c_{n2} & \cdots & 1
\end{bmatrix}^{\left(1\right)}\xrightarrow[2^{\textit{nd}}\,\,\textsl{step}]{\textsl{}}
\]

\[
\xrightarrow[2^{\textit{nd}}\,\,\textsl{step}]{\textsl{}}\begin{bmatrix}1 & c_{12} & \cdots & 1/\alpha\\
c_{21} & 1 &  & 1/\left(\alpha c_{12}\right)\\
\vdots &  & \ddots & \vdots\\
\alpha & \alpha c_{12} & \cdots & 1
\end{bmatrix}^{\left(2\right)}=\boldsymbol{C}^{\left(2\right)}
\]

After finishing the algorithm, we may expect that $\boldsymbol{C}^{\left(m\right)}$
contains $m\leq n-1$ altered elements compared to $\boldsymbol{C}^{\left(0\right)}$.
It may turn out that $m<n$. In order to determine the moment when
the algorithm should be stopped after each step of the calculations,
it is necessary to check the condition $r_{C'}(a_{q})<r_{C'}(a_{p})$.
If it is true, we finish the algorithm.

Let us present the above procedure in the form of a structural pseudo-code
(Code listing \ref{lis:Simple-algorithm,-FindM}). 

\begin{algorithm}[H]
\caption{FindM() method}\label{lis:Simple-algorithm,-FindM}

\begin{algorithmic}[1]

\Function{FindM}{$C$, p, q}

\State  $\alpha$ $\leftarrow$ set to the initial value;\label{FindM:alpha-initiation}

\State  $m^{res}$ $\leftarrow$ $2 \cdot n$, where $n$ is a dimension
of the matrix $C$\label{FindM:mres-init-value}

\State  $C^{res}$ $\leftarrow$ $C$\label{FindM:Cres-initiation}

        \While{($\alpha$ > $1$)}\label{FindM:while-beginning}

\State    $m^{temp}$, $C^{temp}$ $\leftarrow$ ComputeChanges($C$, $\alpha$, p, q)\label{FindM:computeChanges-call}

          \If{$(m^{temp}$ < $m^{res})$}\label{FindM:IfConstraint}

\State       $m^{res}$ $\leftarrow$ $m^{temp}$\label{FindM:mres-assignment}

\State       $C^{res}$ $\leftarrow$ $C^{temp}$\label{FindM:cres-assignment}

          \Else

             \If{$(m^{temp}$ = $m^{res})$ \textbf{and} $(I(C^{temp})$ $\leq$ $I(C^{res}))$}\label{FindM:If2Constraint}

\State           $C^{res}$ $\leftarrow$ $C^{temp}$\label{FindM:cres2-assignment}

             \EndIf

          \EndIf

\State    decrement $\alpha$\label{FindM:alpha-decrement}

        \EndWhile\label{FindM:while-end}

\State\Return$m^{\textsl{res}}$ and $C^{res}$\label{FindM:returned-values}

\EndFunction

\alglinenoNew{alg1}\alglinenoPush{alg1}

\end{algorithmic}
\end{algorithm}

The heart of the algorithm is the transformation \eqref{eq:Matrix C elements dependence}.
Indeed, we can find it in the code of $\textsl{ComputeChanges}\left(\right)$
(Listing: \ref{lis:Simple-algorithm,-ComputeChanges}, lines: \ref{CompChangesSimple:cres-pj-update}
- \ref{CompChangesSimple:cres-jp-update}). In the context of the
definition of inconsistency \eqref{eq:entries_in_C}, the use of transformation
\eqref{eq:Matrix C elements dependence} seems to be quite natural.
Thus, for the purposes of this paper, we will refer to the approach
presented in this section as the row algorithm. The maximal number
of main loop (Listing: \ref{lis:Simple-algorithm,-ComputeChanges},
lines: \ref{CompChangesSimple:for-begin} - \ref{CompChangesSimple:for-end})
iterations is limited by the size of the input matrix $C$, so it
is less or equal to $n-2$, and is strictly correlated with the number
of modified elements ($n-1$). Taking into consideration the above
observations, we could conclude that the\emph{ row algorithm} is useful
especially when the input matrix is large, or the computational resources
are limited. It is worth noting that this algorithm is also resistant
to the inconsistency of the input matrix, because the\emph{ ComputeChanges()
}method (Listing: \ref{lis:Simple-algorithm,-ComputeChanges}) does
not need an $\textit{I}(C)$ value\footnote{$I(C)$ denotes the value of the inconsistency index for $C$ determined
using $\textit{CI},\,\textit{GCI}$ or $\textit{ISLD}$. } in any of its steps. The number of steps is not associated with the
\emph{consistency ratio }value\emph{, }it depends only on the $r_{C}$'s
comparison result (Listing \ref{lis:Simple-algorithm,-FindM}, line:
\ref{CompChangesSimple:break-condition}). Thus, even an inconsistent
matrix will be a valid input for the discussed method.

For the purposes of this study, the \emph{row algorithm} was written
in the form of two functions: \emph{FindM()} (Listing \ref{lis:Simple-algorithm,-FindM})
and \emph{ComputeChanges()} (Listing \ref{lis:Simple-algorithm,-ComputeChanges}).
\emph{FindM()} iterates through different $\alpha$'s and chooses
the best option proposed by \emph{ComputeChanges()}, which is responsible
for updating elements of $\boldsymbol{C}$. \emph{The FindM()} function
is used in both of the described algorithms and its implementation
does not change, unlike the \emph{ComputeChanges()} function implementation,
which is algorithm dependent. The \emph{ComputeChanges()} (Listing
\ref{lis:Simple-algorithm,-ComputeChanges}) method returns the number
of modified elements in the matrix $\boldsymbol{C}$ and modified
matrix $\boldsymbol{C}^{\textsl{res}}$. Those values are compared
inside the \emph{FindM()} function and the best solution is selected.

At the beginning, \emph{FindM()} sets $\alpha$ to the user-defined
initial value (Listing \ref{lis:Simple-algorithm,-FindM}, \ref{FindM:alpha-initiation})
not greater than $1$. Then the variables $m^{\textsl{res}}$ and
$\boldsymbol{C}^{\textsl{res}}$ are initialized (Listing \ref{lis:Simple-algorithm,-FindM},
lines: \ref{FindM:mres-init-value} - \ref{FindM:Cres-initiation}).
Next, the \emph{while} loop is executed as long as $\alpha$ is greater
than $1$ (lines: \ref{FindM:while-beginning} - \ref{FindM:while-end}).
The first operation inside \emph{while} (line: \ref{FindM:IfConstraint})
calls the $\textsl{ComputeChanges}\left(\right)$ method (Listing
\ref{lis:Simple-algorithm,-ComputeChanges}), and stores the returned
results as $m^{\textsl{temp}}$ and $\boldsymbol{C}^{\textsl{temp}}$.
Then, if $\textsl{ComputeChanges}\left(\right)$ alters fewer elements
than it did in the previous step (condition in the line: \ref{FindM:IfConstraint})
$m^{\textsl{res}}$ and $\boldsymbol{C}^{\textsl{res}}$ are correspondingly
updated (lines: \ref{FindM:mres-assignment}, \ref{FindM:cres-assignment}).
If it does not, providing that the number of altered elements in the
current and the previous step is the same, and inconsistency of the
matrix $\boldsymbol{C}$ decreased, $\boldsymbol{C}^{\textsl{res}}$
is replaced by its more consistent version $\boldsymbol{C}^{\textsl{temp}}$.
At the end of the while loop, $\alpha$ is decreased (line: \ref{FindM:alpha-decrement}).
The method ends with returning both: the number of modified elements
and the matrix itself (line: \ref{FindM:returned-values}).

The next part of our row algorithm is contained in $\textsl{ComputeChanges}\left(\right)$
(Listing \ref{lis:Simple-algorithm,-ComputeChanges}). The method
starts with establishing the input parameters for further calculations
(lines: \ref{CompChangesSimple:cres-init-value} - \ref{CompChangesSimple:sort-p-row-init-value}).
Then the for loop is executed (lines: \ref{CompChangesSimple:for-begin}
- \ref{CompChangesSimple:for-end}). It iterates through elements
from $row_{p}^{sorted}$ as long as the rank $a_{q}$ is greater than
the rank $a_{p}$ or any chosen inconsistency index for $\boldsymbol{C}^{\textsl{res}}$
is greater than the appropriate threshold value (line: \ref{CompChangesSimple:break-condition}).
Then, we perform the desired update of $c_{\textsl{p}\textsl{j}}^{\textsl{res}}$
and its reciprocal element $c_{jp}^{\textsl{res}}$ (lines: \ref{CompChangesSimple:cres-pj-update}
and \ref{CompChangesSimple:cres-jp-update}). We also increment the
number of modified elements $m^{\textsl{res}}$. $\textsl{ComputeChanges}\left(\right)$
ends up with returning the results: the number of modifications: $m^{\textsl{res}}$
and the modified matrix $\boldsymbol{C}^{\textsl{res}}$ (line: \ref{CompChangesSimple:returned-values}).
 As the size of $\boldsymbol{C}^{\textsl{res}}$ is finite, the stop
condition of $\textsl{ComputeChanges}\left(\right)$ is met. This
condition is crucial, as it ensures that only the previously unchanged
element is modified in each turn of the loop. Therefore, the running
time of the procedure is $O(n^{2})$.

\begin{algorithm}[H]
\caption{ComputeChanges() method - row algorithm}\label{lis:Simple-algorithm,-ComputeChanges}

\begin{algorithmic}[1]

\alglinenoPop{alg1}

\Function{ComputeChanges}{$C$, $\alpha$, p, q}

\State$C^{res}$ $\leftarrow$ $C$\label{CompChangesSimple:cres-init-value}

\State$c^{res}_{pq}$ $\leftarrow$ $\alpha$\label{CompChangesSimple:cres-pq-init-value}

\State$c^{res}_{qp}$ $\leftarrow$ $1/\alpha$\label{CompChangesSimple:cres-qp-init-value}

\State$m^{res}$ $\leftarrow$ $1$\label{CompChangesSimple:mres-init-value}

\State$row_{p}$ $\leftarrow$ \text{get the $p$-th row of $C^{res}$ and remove elements at position $p$ and $q$}\label{CompChangesSimple:p-row-init-value}

\State$row_{p}^{sorted}$ $\leftarrow$ \text{sort $row_{p}$ elements in ascending order}\label{CompChangesSimple:sort-p-row-init-value}

\For{each subsequent element $c^{res}_{pj}$ in the $row_{p}^{sorted}$}\label{CompChangesSimple:for-begin}

\If{$r_{C^{res}}(a_{p})>r_{C^{res}}(a_{q})$ \textbf{and} $I(C^{res})$ < threshold}\label{CompChangesSimple:break-condition}

\State\textbf{exit loop}\label{CompChangesSimple:break-point}

\Else

\State$c^{res}_{pj}$ $\leftarrow$ $\alpha \cdot c^{res}_{qj}$\label{CompChangesSimple:cres-pj-update}

\State$c^{res}_{jp}$ $\leftarrow$ $1/c^{res}_{pj}$\label{CompChangesSimple:cres-jp-update}

\State\text{increment} $m^{res}$\label{CompChangesSimple:mres-update}

\EndIf

\EndFor\label{CompChangesSimple:for-end}

\State\Return$2 \cdot m^{\textsl{res}}$ and $C^{res}$\label{CompChangesSimple:returned-values}

\EndFunction

\alglinenoPush{alg1}

\end{algorithmic}
\end{algorithm}

Let us see how the PC matrix changes during the execution of $\textsl{ComputeChanges}\left(\right)$
operation (items changed in each iteration are underlined). Let the
change factor $\alpha$ be set to $1.2$, where inconsistency index
\emph{I} is \emph{CI} (\ref{eq:saaty-idx}), and priority vector is
calculated using EVM (\ref{eq:evm-element-val}):

\[
\begin{array}{c}
\underbrace{\boldsymbol{C}^{\left(0\right)}=\begin{bmatrix}1 & 0.3203 & 6.4158 & 1.5449\\
3.1224 & 1 & 3.2254 & 1.7171\\
0.1559 & 0.3100 & 1 & 0.1390\\
0.6473 & 0.5824 & 7.1927 & 1
\end{bmatrix}}\\
\begin{array}{cc}
w^{\text{\ensuremath{\left(0\right)}}}=\begin{bmatrix}0.2657\\
0.4242\\
0.0606\\
0.2495
\end{bmatrix}, & \begin{array}{ll}
\textit{I}\left(\boldsymbol{C}^{\left(0\right)}\right) & =0.1420\\
\textsl{p} & =3\\
\textsl{q} & =2
\end{array}\end{array}
\end{array}\xrightarrow[\textsl{iter.}]{\textsl{1st}}
\]

\[
\xrightarrow[\textsl{iter.}]{\textsl{1st}}\begin{array}{c}
\underbrace{\boldsymbol{C}^{\left(1\right)}=\begin{bmatrix}1 & 0.3203 & 6.4158 & 1.5449\\
3.1224 & 1 & \underline{0.8333} & 1.7171\\
0.1559 & \underline{1.2000} & 1 & 0.1390\\
0.6473 & 0.5824 & 7.1927 & 1
\end{bmatrix}}\\
\begin{array}{cc}
w^{\text{\ensuremath{\left(1\right)}}}=\begin{bmatrix}0.2849\\
0.3332\\
0.1110\\
0.2709
\end{bmatrix}, & \begin{array}{ll}
\textit{I}\left(\boldsymbol{C}^{\left(1\right)}\right) & =0.4477\\
r_{C}(a_{\textit{p}}) & =1\\
r_{C}(a_{\textit{q}}) & =4
\end{array}\end{array}
\end{array}\xrightarrow[\textsl{iter.}]{\textsl{2nd}}
\]

\[
\xrightarrow[\textsl{iter.}]{\textsl{2nd}}\begin{array}{c}
\underbrace{\boldsymbol{C}^{\left(2\right)}=\begin{bmatrix}1 & 0.3203 & 6.4158 & 1.5449\\
3.1224 & 1 & 0.8333 & 1.7171\\
0.1559 & 1.2000 & 1 & \underline{2.0605}\\
0.6473 & 0.5824 & \underline{0.4853} & 1
\end{bmatrix}}\\
\begin{array}{cc}
w^{\text{\ensuremath{\left(2\right)}}}=\begin{bmatrix}0.3482\\
0.3503\\
0.1769\\
0.1246
\end{bmatrix}, & \begin{array}{ll}
\textit{I}\left(\boldsymbol{C}^{\left(2\right)}\right) & =0.3783\\
r_{C}(a_{\textit{p}}) & =2\\
r_{C}(a_{\textit{q}}) & =4
\end{array}\end{array}
\end{array}\xrightarrow[\textsl{iter.}]{\textsl{3rd}}
\]

\[
\xrightarrow[\textsl{iter.}]{\textsl{3rd}}\begin{array}{c}
\underbrace{\boldsymbol{C}^{\left(3\right)}=\begin{bmatrix}1 & 0.3203 & \underline{0.2669} & 1.5449\\
3.1224 & 1 & 0.8333 & 1.7171\\
\underline{3.7469} & 1.2000 & 1 & 2.0605\\
0.6473 & 0.5824 & 0.4853 & 1
\end{bmatrix}}\\
\begin{array}{cc}
w^{\text{\ensuremath{\left(3\right)}}}=\begin{bmatrix}0.1394\\
0.3235\\
0.3882\\
0.1489
\end{bmatrix}, & \begin{array}{ll}
\textit{I}\left(\boldsymbol{C}^{\left(3\right)}\right) & =0.0454\\
r_{C}(a_{\textit{p}}) & =4\\
r_{C}(a_{\textit{q}}) & =3
\end{array}\end{array}
\end{array}
\]

In each iteration $\textsl{ComputeChanges}\left(\right)$ modifies
different elements of $\boldsymbol{C}$. This in turn causes the value
of $w$ and consistency of $C$ to change. In the above example, in
the third iteration it holds that $r_{C^{\left(3\right)}}(a_{\textit{q}})<r_{C^{\left(3\right)}}(a_{\textit{p}})$.
This means the intended goal is reached and the algorithm completed.
It is worth noting that the resulting matrix meets the usual consistency
criterion for \emph{CI} i.e. it holds that $I\left(\boldsymbol{C}^{\left(3\right)}\right)\leq0.1$.

\section{Matrix algorithm\label{sec:Matrix-algorithm}}

In this section, we would like to introduce a slightly different approach
to finding appropriate modifications. The priority vector $w$ is
directly computed from $\boldsymbol{C}$\textbf{ }(equation \ref{eq:eigenvalueeq})
and if matrix $\boldsymbol{C}$ is consistent, each $c_{ij}$ element
is equal to the ratio of ranking results for $a_{i}$ and $a_{j}$
(equation \ref{eq:entries_in_C}). Hence, the consistent matrix $\boldsymbol{C}^{con}$
can be computed directly from $w$.

Let us also introduce a way to measure the distance between elements
of two distinctive matrices $\boldsymbol{A}$ and $\boldsymbol{B}$.
The best candidate is a Hadamard product of matrices $\boldsymbol{A}$
and $\boldsymbol{B}^{T}$ defined as:

\begin{equation}
\boldsymbol{H}=[a_{ij}\cdot b_{ji}]=\boldsymbol{A}\bullet\boldsymbol{B}^{T}\label{eq:Hadamard product}
\end{equation}

\noindent which indicates that if $\left|a_{ij}-b_{ij}\right|\thickapprox0$
then the value of $h_{ij}$ and $h_{ji}$ tends to $1$. It is worth
noting that the value of one element is greater than $1$, while the
other one is lower.

\noindent The main idea of this algorithm is to change elements of
matrix $C$ - $c_{ij}$ to its counterparts - $c_{ij}^{con}$ starting
from the most distant ones to the closest. Distances are computed
using (equation \ref{eq:Hadamard product}), and only the ones which
are greater than $1$ are taken into consideration. For the specific
element $c_{ij}$ its reciprocal element $c_{ji}$ is also changed.
Such a way of choosing and modifying elements leads to the conclusion
that the maximal number of modified elements is:

\begin{equation}
m_{max}^{res}=n(n-1)\label{eq:IIIalg, max number of modified elements}
\end{equation}

\noindent where $n$ is the size of the matrix $C$. The goal of all
algorithms remains the same, making $\textit{rank}_{C^{'}}(a_{\textit{fp}})\leq\textit{rank}_{C^{'}}(a_{\textit{ip}})$,
so the final value of $m^{res}$ might be lower than $m_{max}^{res}$.

\noindent The idea of the \emph{matrix algorithm} is presented in
the listing \ref{lis:IIIalgorithm,ComputeChanges}. All of the input
parameters are the same as in the \emph{row algorithm}. In the first
step, we initialize local variables (lines: \ref{CompChangesIII:Cres initial value},
\ref{CompChangesIII:mres initial value}) and compute important helper
variables (lines: \ref{CompChangesIII:calculate w for C}-\ref{CompChangesIII:generate respos vector}).
To compute $C^{con}$ (line: \ref{CompChangesIII:generate consistent matrix})
we use a modified priority vector $w'$ which promotes alternative
$a_{p}$ over $a_{q}$. Then the for loop is executed (lines: \ref{CompChangesIII:for begin}-\ref{CompChangesIII:for end}).
It iterates through elements from $respos$ as long as the rank of
$a_{q}$ is greater than the rank of $a_{p}$ or any chosen inconsistency
index for $\boldsymbol{C}^{\textsl{res}}$ is greater than some threshold
value (line: \ref{CompChangesIII:break condition}), otherwise the
loop is exited (line: \ref{CompChangesIII:break point}). Then, we
perform the desired update of $c_{ij}^{res}$ (line: \ref{CompChangesIII:cres_ij update}),
its reciprocal element $c_{ji}^{res}$ (line: \ref{CompChangesIII:cres_ji update})
and we increment $m^{res}$ (line: \ref{CompChangesIII:mres update}).
The described method ends up with returning the number of modified
elements - $m^{res}$, and modified matrix $C^{res}$ (line: \ref{CompChangesIII:return values}).
Since in every turn of the loop previously unchanged elements are
modified, after the last possible turn $C^{res}$ is equal to $C^{con}$
and the stop condition is met. For the same reason, the running time
of the procedure is $O(n^{2})$.
\begin{algorithm}[H]
\caption{ComputeChanges() method - matrix algorithm}\label{lis:IIIalgorithm,ComputeChanges}

\begin{algorithmic}[1]

\alglinenoNew{alg3}

\Function{ComputeChanges}{$C$, $\alpha$, p, q}\label{CompChangesIII:input parameters}

\State$C^{res}$ $\leftarrow$ $C$\label{CompChangesIII:Cres initial value}

\State$m^{res}$ $\leftarrow$ $0$\label{CompChangesIII:mres initial value}

\State$w$ $\leftarrow$ \text{calculate priority vector for matrix $C$}\label{CompChangesIII:calculate w for C}

\State$C^{con}$ $\leftarrow$ \text{generate consistent matrix using vector $w$, where $w_{p} = \alpha w_{q}$}\label{CompChangesIII:generate consistent matrix}

\State$C^{had}$ $\leftarrow$ \text{calculate Hadamard product for matrix $C$ and matrix ${C^{con}}^T$}\label{CompChangesIII:calculate hadamard matrix}

\State$respos$ $\leftarrow$ \text{sort $C^{had}$'s elements, greater than one, in descending order}\label{CompChangesIII:generate respos vector}

\For{each subsequent element $c^{had}_{ij}$ in the $respos$}\label{CompChangesIII:for begin}

\If{$r_{C^{res}}(a_{p})>r_{C^{res}}(a_{q})$ \textbf{and} $I(C^{res})$ < threshold}\label{CompChangesIII:break condition}

\State\textbf{exit loop}\label{CompChangesIII:break point}

\Else

\State$c^{res}_{ij}$ $\leftarrow$ $c^{con}_{ij}$\label{CompChangesIII:cres_ij update}

\State$c^{res}_{ji}$ $\leftarrow$ $c^{con}_{ji}$\label{CompChangesIII:cres_ji update}

\State\text{increment} $m^{res}$\label{CompChangesIII:mres update}

\EndIf

\EndFor\label{CompChangesIII:for end}

\State\Return$2 \cdot m^{\textsl{res}}$ and $C^{res}$\label{CompChangesIII:return values}

\EndFunction

\alglinenoPush{alg3}

\end{algorithmic}
\end{algorithm}

The workflow of the $\textsl{ComputeChanges}\left(\right)$ method
(Listing \ref{lis:IIIalgorithm,ComputeChanges}) is presented in the
example below. Let the change parameter $\alpha$ be the same as previously
i.e. ($\alpha=1.2$), where the priority vector deriving method is
EVM (\ref{eq:evm-element-val}) and the inconsistency index \emph{I}
is \emph{CI} (\ref{eq:saaty-idx}). The initial values of $C^{res}$
(Listing \ref{lis:IIIalgorithm,ComputeChanges}, line: \ref{CompChangesIII:Cres initial value}),
$C^{had}$ (line: \ref{CompChangesIII:calculate hadamard matrix})
and $\textit{respos}$ (line: \ref{CompChangesIII:generate respos vector})
are as follows:

\medskip{}

$C^{res}=\begin{bmatrix}1 & 0.3203 & 6.4158 & 1.5449\\
3.1224 & 1 & 3.2254 & 1.7171\\
0.1559 & 0.3100 & 1 & 0.1390\\
0.6473 & 0.5824 & 7.1927 & 1
\end{bmatrix}$

\medskip{}

$C^{had}=\begin{bmatrix}1 & 0.51148 & 12.2943 & 1.45109\\
1.95532 & 1 & 3.87048 & 1.00999\\
0.0813567 & 0.258333 & 1 & 0.0681328\\
0.689146 & 0.990145 & 14.6741 & 1
\end{bmatrix}$

\medskip{}

$\textit{respos}=\left\{ 14.6741,\,12.2943,\,3.87048,\,1.95532,\,1.45109,\,1.00999\right\} $

\medskip{}

\noindent consecutive steps of the described algorithm are presented
below, as well as a detailed description of important variables at
each step:

\[
\begin{array}{c}
\underbrace{C^{\left(0\right)}=\begin{bmatrix}1 & 0.3203 & 6.4158 & 1.5449\\
3.1224 & 1 & 3.2254 & 1.7171\\
0.1559 & 0.3100 & 1 & 0.1390\\
0.6473 & 0.5824 & 7.1927 & 1
\end{bmatrix}}\\
\begin{array}{cc}
w^{\text{\ensuremath{\left(0\right)}}}=\begin{bmatrix}0.2657\\
0.4242\\
0.0606\\
0.2495
\end{bmatrix}, & \begin{array}{ll}
\textit{I}\left(\boldsymbol{C}^{\left(0\right)}\right) & =0.1420\\
\textsl{p} & =3\\
\textsl{q} & =2
\end{array}\end{array}
\end{array}\xrightarrow[\textsl{iter.}]{\textsl{1st}}
\]

\[
\xrightarrow[\textsl{iter.}]{\textsl{1st}}\begin{array}{c}
\underbrace{C^{\left(1\right)}=\begin{bmatrix}1 & 0.3203 & 6.4158 & 1.5449\\
3.1224 & 1 & 3.2254 & 1.7171\\
0.1559 & 0.3100 & 1 & \underline{2.0401}\\
0.6473 & 0.5824 & \underline{0.4902} & 1
\end{bmatrix}}\\
\begin{array}{cc}
w^{\text{\ensuremath{\left(1\right)}}}=\begin{bmatrix}0.3084\\
0.4301\\
0.1242\\
0.1374
\end{bmatrix}, & \begin{array}{ll}
\textit{I}\left(\boldsymbol{C}^{\left(1\right)}\right) & =0.2395\\
r_{C}(a_{\textit{p}}) & =1\\
r_{C}(a_{\textit{q}}) & =4
\end{array}\end{array}
\end{array}\xrightarrow[\textsl{iter.}]{\textsl{2nd}}
\]

\[
\xrightarrow[\textsl{iter.}]{\textsl{2nd}}\begin{array}{c}
\underbrace{C^{\left(2\right)}=\begin{bmatrix}1 & 0.3203 & \underline{0.5219} & 1.5449\\
3.1224 & 1 & 3.2254 & 1.7171\\
\underline{1.9163} & 0.3100 & 1 & 2.0401\\
0.6473 & 0.5824 & 0.4902 & 1
\end{bmatrix}}\\
\begin{array}{cc}
w^{\text{\ensuremath{\left(2\right)}}}=\begin{bmatrix}0.1554\\
0.4626\\
0.2319\\
0.1501
\end{bmatrix}, & \begin{array}{ll}
\textit{I}\left(\boldsymbol{C}^{\left(2\right)}\right) & =0.0749\\
r_{C}(a_{\textit{p}}) & =3\\
r_{C}(a_{\textit{q}}) & =4
\end{array}\end{array}
\end{array}\xrightarrow[\textsl{iter.}]{\textsl{3rd}}
\]

\[
\xrightarrow[\textsl{iter.}]{\textsl{3rd}}\begin{array}{c}
\underbrace{C^{\left(3\right)}=\begin{bmatrix}1 & 0.3203 & 0.5219 & 1.5449\\
3.1224 & 1 & \underline{0.8333} & 1.7171\\
1.9163 & \underline{1.2} & 1 & 2.0401\\
0.6473 & 0.5824 & 0.4902 & 1
\end{bmatrix}}\\
\begin{array}{cc}
w^{\text{\ensuremath{\left(3\right)}}}=\begin{bmatrix}0.1678\\
0.3433\\
0.3364\\
0.1525
\end{bmatrix}, & \begin{array}{ll}
\textit{I}\left(\boldsymbol{C}^{\left(3\right)}\right) & =0.0351\\
r_{C}(a_{\textit{p}}) & =3\\
r_{C}(a_{\textit{q}}) & =4
\end{array}\end{array}
\end{array}\xrightarrow[\textsl{iter.}]{\textsl{4th}}
\]

\[
\xrightarrow[\textsl{iter.}]{\textsl{4th}}\begin{array}{c}
\underbrace{C^{\left(4\right)}=\begin{bmatrix}1 & \underline{0.6262} & 0.5219 & 1.5449\\
\underline{1.5969} & 1 & 0.8333 & 1.7171\\
1.9163 & 1.2 & 1 & 2.0401\\
0.6473 & 0.5824 & 0.4902 & 1
\end{bmatrix}}\\
\begin{array}{cc}
w^{\text{\ensuremath{\left(4\right)}}}=\begin{bmatrix}0.2013\\
0.2923\\
0.3499\\
0.1565
\end{bmatrix}, & \begin{array}{ll}
\textit{I}\left(\boldsymbol{C}^{\left(4\right)}\right) & =0.0056\\
r_{C}(a_{\textit{p}}) & =4\\
r_{C}(a_{\textit{q}}) & =3
\end{array}\end{array}
\end{array}.
\]

In each iteration, the PC matrix \textbf{$\boldsymbol{C}$} elements
are modified. Modified elements are underlined to clearly present
the workflow of the algorithm. Because it is known in advance how
many elements might be modified (equation \ref{eq:IIIalg, max number of modified elements}),
we might expect that the value of $m^{res}$ is greater than in previously
described algorithms (Listing \ref{lis:Simple-algorithm,-ComputeChanges}).
Indeed, the total number of modified matrix $\boldsymbol{C}$ elements
is $8$, while ``row'' algorithms modified only $6$ elements. After
execution of the $\textsl{ComputeChanges}\left(\right)$ method (Listing
\ref{lis:IIIalgorithm,ComputeChanges}) $m^{\textsl{res}}$ and $\boldsymbol{C}^{\textsl{res}}$
are returned inside the slightly modified $\textsl{FindM}\left(\right)$
method (Listing \ref{lis:Simple-algorithm,-FindM}). This modification
is a change of the initial $m^{res}$ value from $2n$ to $n^{2}$
(line: \ref{FindM:mres-init-value}). To find possibly better solutions,
the $\textsl{ComputeChanges}\left(\right)$ method might be executed
again with different $\alpha$ values.

\section{Construction of two algorithms - a comparison\label{sec:Construction-comparison}}

The first algorithm (Sec. \ref{sec:Row-algorithm}) performs fewer
steps and is easier to implement than the second one (Sec. \ref{sec:Matrix-algorithm}).
The main advantage of the $\textsl{ComputeChanges}\left(\right)$
method (Listing \ref{lis:Simple-algorithm,-ComputeChanges}) over
its counterpart (Listing \ref{lis:IIIalgorithm,ComputeChanges}) is
that the number of modified elements in the PC matrix $C$ is limited
to $2(n-1)$, where $n$ is the size of $C$, over $n\left(n-1\right)$.
The disadvantage of the ``row'' $\textsl{ComputeChanges}\left(\right)$
method (Listing \ref{lis:Simple-algorithm,-ComputeChanges}) is it
minimizes inconsistency in a slightly chaotic way. Thus, the matrix
algorithm's $\textsl{ComputeChanges}\left(\right)$ (Listing \ref{lis:IIIalgorithm,ComputeChanges})
not only gives better results, but also it tries to minimize inconsistency
in each step where it is possible. Such a strategy is more effective
but, of course, requires more computing power. Both algorithms stop
after a finite number of operations are performed. The first\emph{
}$\textsl{ComputeChanges}\left(\right)$ method (Listing \ref{lis:Simple-algorithm,-ComputeChanges})
has the asymptotic running time of $O(n)$, unlike its counterpart
for the matrix algorithm (Listing \ref{lis:IIIalgorithm,ComputeChanges})
which has the computational complexity of $O(n^{2})$.

Both algorithms correctly changed initial matrix elements for the
same initial PC matrix $\boldsymbol{C}$. The alternative, expected
to be ranked as the best, was successfully promoted. To simplify the
workflow of the algorithms (Sec. \ref{sec:Row-algorithm}, Sec. \ref{sec:Matrix-algorithm}),
we decided to summarize their outcome shortly. The input matrix $\boldsymbol{C}$
and the output matrix $\boldsymbol{C}^{res}$ for the\emph{ row algorithm}
are presented below:

\[
\boldsymbol{C}=\begin{bmatrix}1 & 0.320 & 6.416 & 1.545\\
3.122 & 1 & 3.225 & 1.717\\
0.156 & 0.310 & 1 & 0.139\\
0.647 & 0.582 & 7.193 & 1
\end{bmatrix}\longrightarrow\boldsymbol{C}^{res}=\begin{bmatrix}1 & 0.320 & 0.267 & 1.545\\
3.122 & 1 & 0.833 & 1.717\\
3.747 & 1.200 & 1 & 2.060\\
0.647 & 0.582 & 0.485 & 1
\end{bmatrix},
\]
and similarly, $\boldsymbol{C}$ \& $\boldsymbol{C}^{res}$ for the\emph{
matrix algorithm}:
\[
\boldsymbol{C}=\begin{bmatrix}1 & 0.320 & 6.416 & 1.545\\
3.122 & 1 & 3.225 & 1.717\\
0.156 & 0.310 & 1 & 0.139\\
0.647 & 0.582 & 7.193 & 1
\end{bmatrix}\longrightarrow\boldsymbol{C}^{res}=\begin{bmatrix}1 & 0.626 & 0.522 & 1.545\\
1.597 & 1 & 0.833 & 1.717\\
1.916 & 1.2 & 1 & 2.040\\
0.647 & 0.582 & 0.490 & 1
\end{bmatrix}.
\]

The \emph{row algorithm} changed fewer elements in the input matrix
$\boldsymbol{C},$ but the value of the inconsistency index for the
output matrix $\boldsymbol{C}^{res}$ is higher. On the other hand,
the \emph{matrix algorithm }executes more steps before returning the
correct results, but the value of the inconsistency index is smaller
by orders of magnitude.

\section{Monte Carlo experiments\label{sec:Monte-Carlo-experiments}}

To verify the proposed algorithms in practice, we conducted a number
of Monte Carlo experiments. We paid special attention to the following:
\begin{itemize}
\item range of the initial inconsistency in which algorithms give acceptable
results
\item the number of changes to individual elements needed to perform manipulations
\end{itemize}
Knowing these two values in practice will allow us to assess the effectiveness
of the attack (threat scales) with the use of the proposed heuristics.

For the purposes of the tests, we generate sets of $1000$ inconsistent
PC matrices for which the inconsistency index\footnote{Although we used the CI index to prepare the matrices, the Montecarlo
experiments were also carried out for other inconsistency indices.} \emph{CI} fits the range $[0.005k,\,0.005\left(k+1\right)]$, where
$k=0,1,2\ldots$. To create a single $n\times n$ inconsistent matrix,
we use the disturbance factor $d$. The generation process relies
on creating a real random ranking $w$ with values between $1/15$
and $15$. Then, we create the consistent matrix $C$ assuming $c_{ij}=w(a_{i})/w(a_{j})$.
To make the matrix inconsistent, we disturb every entry multiplying
it by a randomly chosen number from the range $[1/d,\,d]$. If the
$\textit{CR}$ of such a disturbed PC matrix is smaller than some
acceptance threshold (the default value is $0.1$) we add that matrix
into the testing set $\mathcal{C}^{\textit{in}}$. Otherwise, we keep
generating the next matrices until we get $1000$ items in the testing
set. 

The outcome of the algorithm form a set of result matrices $\mathcal{C}^{\textit{res}}$,
i.e. the output (Sec. \ref{sec:Row-algorithm} and \ref{sec:Matrix-algorithm})
is an appropriately manipulated matrix. To be considered correct,
it should not be too inconsistent. Therefore, we require the input
and the \textquotedbl revised\textquotedbl{} matrix to have inconsistency
not greater than some threshold value $\gamma$. For this purpose,
we used three inconsistency indices, \emph{CI} (\ref{eq:saaty-idx}),
\emph{GCI} (\ref{eq:gci-idx}), and \emph{ISLD} (\ref{eq:kazib-idx}).
As a rule of thumb, we took the threshold of acceptability for the
\emph{CI} index to be $0.1$ \citep{Saaty1977asmf}. We experimentally
determined the thresholds of acceptability for the other \emph{GCI}
and \emph{ISLD} indices. I.e., we generated $1.000$ matrices with
\emph{CI} inconsistency close to $0.1$ and checked the average inconsistency
of these matrices using \emph{GCI} and \emph{ISLD} indicies. In this
way, we obtained $\gamma$ acceptability threshold values for \emph{GCI}
and \emph{ISLD} of $0.27$ and $0.0065$, respectively. However, we
want to refrain from entering into a discussion on the question of
what the threshold of acceptability is or should be. For us, in the
case of this experiment, the acceptability threshold is just a specific
numerical value above which the risk of questioning the ranking results
may dangerously increase, thus thwarting the intended manipulation.

Since both algorithms are heuristic (based on the heuristic of monotonicity
of the ranking function), it is possible that the algorithm does not
find a satisfactory solution. For the purposes of the Monte Carlo
experiments, we will call the ratio of correct (consistent enough)
outputs to all considered inputs $\textit{SR}$ (success rate):

\[
\textit{SR}\overset{\textit{df}}{=}\frac{\left|\{C^{\textit{res}}\in\mathcal{C}^{\textit{res}}\,:\,\textit{I}(C^{\textit{res}})\leq\gamma\}\right|}{\left|\mathcal{C}^{\textit{res}}\right|},
\]

where $I(C)$ is the inconsistency estimated using \emph{CI}, \emph{GCI}
or \emph{ISLD} and $\gamma$ is an experimentally determined acceptable
threshold value. The second important parameter of the experiments
is the preferential distance between swapped alternatives. Of course,
the bigger it is the more demanding the task, and the more difficult
it is to manipulate the input PC matrix. Let $\Delta_{pq}$ denote
the distance between the position of $a_{p}$ and $a_{q}$ in the
ranking:

\[
\varDelta_{pq}\coloneqq\left|r_{C}\left(a_{\textit{q}}\right)-r_{C}\left(a_{\textit{p}}\right)\right|.
\]

The figures below present the Monte Carlo experiments showing the
relationship between\footnote{Although we conducted the experiments using three different inconsistency
indices, \emph{CI}, \emph{GCI}, and \emph{ISLD}, the results were
similar. Thus, for the brevity of the presentation, the figures presented
were generated using the \emph{CI} index only.} $\textit{SR}$ and $\textit{I}$ for different values of $\Delta_{pq}$.

{\setlength\tabcolsep{0pt}
\begin{figure}[h]
\begin{tabular}{>{\centering}m{0.5\linewidth}>{\centering}m{0.5\linewidth}}
\subfloat[$n=7$, $\varDelta_{pq}=1$]{\includegraphics[width=1\linewidth]{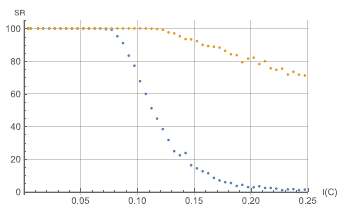}}\\
\subfloat[$n=7$, $\varDelta_{pq}=6$]{\includegraphics[width=1\linewidth]{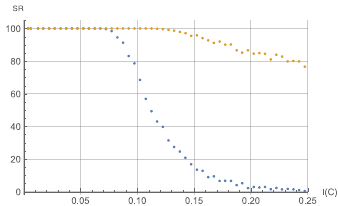}} & \subfloat[$n=9$, $\varDelta_{pq}=1$]{\includegraphics[width=1\linewidth]{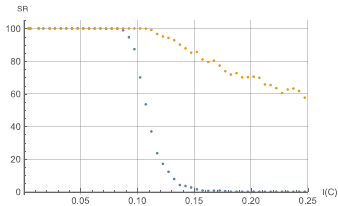}}\\
\subfloat[$n=9$, $\varDelta_{pq}=8$]{\includegraphics[width=1\linewidth]{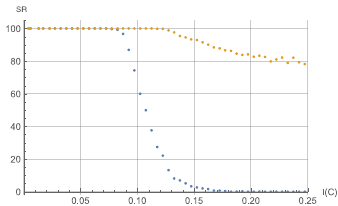}}\tabularnewline
\end{tabular}

\medskip{}

\noindent \begin{centering}
\includegraphics[viewport=0bp -1.5bp 4bp 2.5bp]{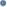}
row algorithm, \includegraphics[viewport=0bp -1.5bp 4bp 2.5bp]{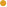}
matrix algorithm
\par\end{centering}
\caption{Success rate for manipulation algorithms with different size of PC
matrices and different preferential distance $\Delta_{pq}$. \label{fig:Change-of-validyty-rate-factor-1}}
\end{figure}
}

It is easy to observe that the \emph{matrix algorithm} performs better
than \emph{the row algorithm}. Indeed, in the case of the former,
the $\textit{SR}$ values decrease slower (Figure \ref{fig:Change-of-validyty-rate-factor-1}).
The \emph{matrix algorithm} seems to be resistant to changes in the
matrix size. The \emph{row algorithm} is the opposite. The larger
the matrix, the more difficult it is to find a sufficiently consistent
solution. An interesting effect of the distance $\Delta pq$ on the
results can be observed in the case of \emph{the matrix algorithm}.
The greater the distance between alternatives subject to swap seems
to be more favorable for it. I.e. in such a case more experiments
are successful.

{\setlength\tabcolsep{0pt}
\begin{figure}[h]
\begin{tabular}{>{\centering}m{0.5\linewidth}>{\centering}m{0.5\linewidth}}
\subfloat[$\varDelta_{pq}=1$]{\includegraphics[width=1\linewidth]{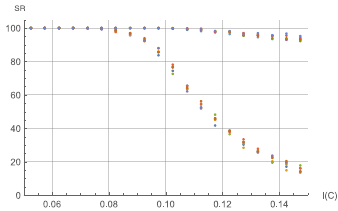}} & \subfloat[$\varDelta_{pq}=5$]{\includegraphics[width=1\linewidth]{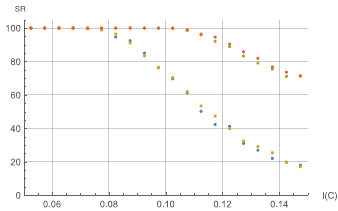}}\tabularnewline
\end{tabular}

\caption{Success rate for $7\times7$ random PC matrices and different $\varDelta_{pq}$.\label{fig:Change-of-the_validity_rate_factor_for_different_deltas_and_sizes}}
\end{figure}
}

The next series of experiments (Fig. \ref{fig:Change-of-the_validity_rate_factor_for_different_deltas_and_sizes})
suggests that $\textit{SR}$ does not depend on $\Delta_{pq}$. This
result is somewhat surprising, as it means that the preferential distance
between the alternatives to be replaced is not a factor decisive for
the success of the manipulation.

We test both algorithms against two ranking methods - EVM and GMM.
The results are quite similar (Fig. \ref{fig:Mean-value-of_m_res-1}),
which seems to support the observation that the lack of monotony of
the former is not that important in practice and does not affect the
operation of algorithms.

{\setlength\tabcolsep{0pt}
\begin{figure}[h]
\begin{tabular}{>{\centering}m{0.5\linewidth}>{\centering}m{0.5\linewidth}}
\subfloat[Input matrix of size $n=5$.]{\includegraphics[width=1\linewidth]{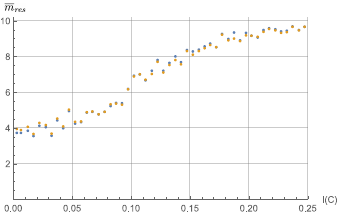}} & \subfloat[Input matrix of size $n=7$.]{\includegraphics[width=1\linewidth]{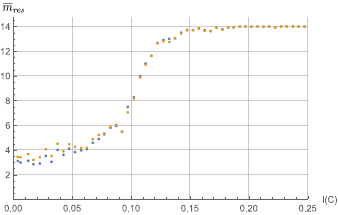}}\tabularnewline
\end{tabular}

\medskip{}

\noindent \begin{centering}
\includegraphics[viewport=0bp -1.5bp 4bp 2.5bp]{blue_dot}
EVM, \includegraphics[viewport=0bp -1.5bp 4bp 2.5bp]{orange_dot}GMM
\par\end{centering}
\caption{Number of matrix elements changed using different priority deriving
methods: EVM and GMM. \label{fig:Mean-value-of_m_res-1}}
\end{figure}
}

As we mentioned earlier, both algorithms were tested using three different
inconsistency indices: \emph{CI}, \emph{GCI} and \emph{ISLD} (Fig.
\ref{fig:Inconsistency_gathering_methods}). 

In our tests, the number $m_{res}$ of altered elements was identical
in all three cases (in Figure \ref{fig:Inconsistency_gathering_methods}b,
the points corresponding to all three inconsistency indices overlap).
Described algorithms use inconsistency indices in sub-sequential steps
(to compare $C^{\textit{res}}$ against the threshold value and to
detect the best result).  Although the inconsistency indices differed,
the general picture of the inconsistency distribution of the tested
matrices (and the various indices) was similar. Thus, in Figure \ref{fig:Inconsistency_gathering_methods}b,
for the purpose of comparing the inconsistency calculated with different
methods, we divided the \emph{GCI} and \emph{ISDL} indices by the
appropriate values of the $\gamma$ coefficient multiplied by $10$.

{\setlength\tabcolsep{0pt}
\begin{figure}[h]
\begin{tabular}{>{\centering}m{0.5\linewidth}>{\centering}m{0.5\linewidth}}
\subfloat[Scaled inconsistency index]{\includegraphics[width=1\linewidth]{IndicesCompScaledValue.pdf}} & \subfloat[$m_{res}$]{\includegraphics[width=1\linewidth]{IndicesCompM_res.pdf}}\tabularnewline
\end{tabular}

\medskip{}

\noindent \begin{centering}
\includegraphics[viewport=0bp -1.5bp 4bp 2.5bp]{blue_dot}
CI, \includegraphics[viewport=0bp -1.5bp 4bp 2.5bp]{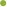}
ISDL, \includegraphics[viewport=0bp -1.5bp 4bp 2.5bp]{orange_dot}
GCI
\par\end{centering}
\caption{Comparison of different inconsistency index gathering methods for
$7\times7$ random PC matrices \label{fig:Inconsistency_gathering_methods}}
\end{figure}
}

The purpose of the last experiment is to check how many elements are
changed by both algorithms. 

For each of the $1000$ matrices (for which the inconsistency $\textit{CI}$
is in range ($[0.005k,\,0.005\left(k+1\right)]$, where $k=0,1,2\ldots$)
we execute both algorithms for every inconsistency index and count
the number of elements changed. Then we calculate the average number
of modified elements $\overline{m}_{res}$ for each of the intervals.
The results are presented below (Fig. \ref{fig:Mean-value-of_m_res}).
Again, due to the similarity of the results, the figures presented
include graphs generated based on the \emph{CI} index only. 

{\setlength\tabcolsep{0pt}
\begin{figure}[h]
\begin{tabular}{>{\centering}m{0.5\linewidth}>{\centering}m{0.5\linewidth}}
\subfloat[PC matrix of size $n=5$.]{\includegraphics[width=1\linewidth]{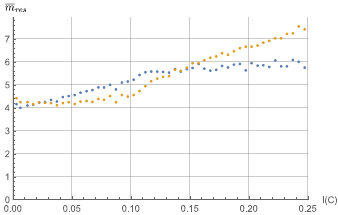}} & \subfloat[PC matrix of size $n=9$.]{\includegraphics[width=1\linewidth]{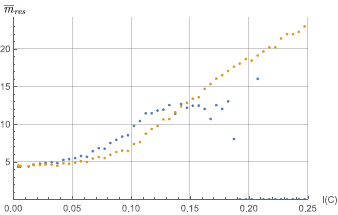}}\tabularnewline
\end{tabular}

\medskip{}

\noindent \begin{centering}
\includegraphics[viewport=0bp -1.5bp 4bp 2.5bp]{blue_dot}
Row algorithm, \includegraphics[viewport=0bp -1.5bp 4bp 2.5bp]{orange_dot}
Matrix algorithm
\par\end{centering}
\caption{Average number of changed elements $\overline{m}_{res}$ for a matrix
with a given inconsistency\label{fig:Mean-value-of_m_res}}
\end{figure}
}

\section{Discussion}

The performed experiments (Sec. \ref{sec:Monte-Carlo-experiments})
show that the algorithms effectively work when some initial conditions
for $C$ are met. One of them is the initial inconsistency of the
input. Let us look at the plots showing the relationship between the
inconsistency and the average number of changed elements $\overline{m}_{res}$
and the success rate $\textit{SR}$ (Fig. \ref{fig:validyty-rate-factor_r_res_comparision}).

{\setlength\tabcolsep{0pt}
\begin{figure}[H]
\begin{tabular}{>{\centering}m{0.5\linewidth}>{\centering}m{0.5\linewidth}}
\subfloat[$\textit{SR}$ and $\overline{m}_{res}$ for PC matrix of size $n=5$]{%
\begin{tabular}{c}
\includegraphics[width=1\linewidth]{5x5MResComparep1q5}\tabularnewline
\includegraphics[width=1\linewidth]{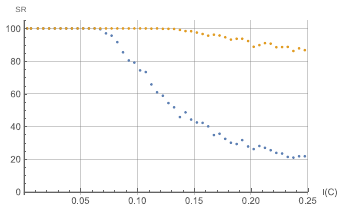}\tabularnewline
\end{tabular}

} & \subfloat[$\textit{SR}$ and $\overline{m}_{res}$ for PC matrix of size $n=9$]{%
\begin{tabular}{c}
\includegraphics[width=1\linewidth]{9x9MResComparep1q9}\tabularnewline
\includegraphics[width=1\linewidth]{9x9p1q9}\tabularnewline
\end{tabular}

}\tabularnewline
\end{tabular}

\medskip{}

\noindent \begin{centering}
\includegraphics[viewport=0bp -1.5bp 4bp 2.5bp]{blue_dot}
Row algorithm, \includegraphics[viewport=0bp -1.5bp 4bp 2.5bp]{orange_dot}
Matrix algorithm
\par\end{centering}
\caption{Comparison of $\overline{m}_{res}$ and $\textit{SR}$ in relation
to the inconsistency index $\textit{CI}$. The results obtained for
the other indices are similar. \label{fig:validyty-rate-factor_r_res_comparision}}
\end{figure}
}

One may observe that the matrix algorithm modified fewer elements
than the row algorithm when inconsistency was not larger than 0.15.
At the same time, \emph{SR} for the matrix algorithm is close to one.
The row algorithm performs worse than the matrix one. For inconsistency
higher than 0.8, its success ratio visibly drops. This result suggests
that the matrix algorithm, albeit a bit more complex, is better than
the row approach.

In the context of an algorithmic attack on the decision-making process,
a question can be asked about the possibility of its detection. In
the case of the row algorithm, following its mode of operation, we
would suggest the following detection method:
\begin{enumerate}
\item Divide every element in the $i$-th row by its counterpart in the
$j$-th row: $\frac{c_{ik}}{c_{jk}}$ for each $i,j,k\in\{1,\ldots,n\}$,
$i\neq j$.
\item If any divided elements have the same value $\frac{c_{ik}}{c_{jk}}=\frac{c_{il}}{c_{jl}}>1$,
it may mean that the i-th alternative has been artificially promoted
using this algorithm.\footnote{Only in cases when more than two elements in the $i$-th row have
been modified.}
\end{enumerate}
The method of detecting the use of the matrix algorithm seems to be
more complicated and cannot be detected that easily. Detection of
the matrix approach will be the subject of research using machine
learning methods.

Both described algorithms work as expected, but their design has several
limitations. No user-defined upper limit should be lower or equal
to the number of possible modifications. This number differs for each
algorithm (Sec. \ref{sec:Construction-comparison}) upper limit for
changed PC matrix $\mathbf{C}$ elements. This feature might be helpful
in some cases, especially when the value of the inconsistency index
is not the main factor and, instead, the more important is the number
of modified elements. We may expect that the inconsistency index might
increase, let's say, by a $10\%$, as long as the number of modified
elements remains lower. When the number of modified elements and the
value of the inconsistency index are somehow correlated, such a case
is worth further investigation.

The value of inconsistency indices for initial input matrices might
be higher than $\gamma$, and even after changes, that value might
be higher than $\gamma$. Such a case sometimes occurs in the row
algorithm; it is a limitation because not all $\mathbf{C}$ matrices
are \textquotedbl manipulable.\textquotedbl{} We do not consider
it a problem because, in such a case, the expert should re-estimate
the value of the alternatives.

Monte Carlo experiments prove that the manipulation of two alternatives
is possible and, in many cases, relatively easy to perform. The necessary
condition (and at the same time, a major limitation of this approach)
to attack the pairwise comparisons decision-making process is the
a priori knowledge of the correct (non-manipulated) input matrix,
so that it is possible to run the manipulation algorithm and enter
its results as the final values of the PC matrix. This assumes either
access of a grafter to an expert (or experts) during the decision-making
process or such organization of the process that the expert has enough
time to carry out calculations over the presented solutions. The above
consideration leads to the observation that, in the case of the threat
of an attack with micro-bribes, the easiest way to protect the decision
procedure is to modify the decision-making process to exclude the
possibility of third-party interaction with the experts. In addition,
it would be worth considering providing experts with appropriate working
conditions with not too much and not too little time for individual
decisions.

\section{Summary}

Decision-making methods based on quantitative pairwise comparisons
are, unfortunately, prone to manipulation. We proposed the manipulation
model and two attack algorithms using the micro-bribes approach in
the presented work. Furthermore, we have analyzed their effectiveness
and indicated the circumstances they can be effective in. In our considerations,
we took into account the issue of data inconsistency. An important
issue that challenges researchers is the detection of manipulation.
In future research, we intend to focus on this issue, especially in
the context of algorithmic manipulations. An important issue that
has not been appropriately addressed is manipulation detection. Thus,
this matter will be a challenge for us in future research.

\section*{Acknowledgments}

The research is supported by The National Science Centre, Poland,
project SODA no. 2021/41/B/HS4/03475. 

This research was funded in whole or in part by National Science Centre,
Poland 2021/41/B/HS4/03475. For the purpose of Open Access, the author
has applied a CC-BY public copyright licence to any Author Accepted
Manuscript (AAM) version arising from this submission.

\bibliographystyle{plain}
\bibliography{papers_biblio_reviewed}

\end{document}